\documentclass[11pt,letterpaper]{article}
\usepackage[margin=0.85in,top=0.8in,bottom=0.85in]{geometry}
\usepackage[T1]{fontenc}
\usepackage{lmodern,amsmath,amssymb,booktabs,longtable,array,graphicx,float,caption,xcolor}
\usepackage[numbers,sort&compress]{natbib}
\usepackage[hidelinks]{hyperref}
\newcommand{\doi}[1]{\url{https://doi.org/#1}}

\hypersetup{pdfauthor={Zeen Song}}
\newcommand{\ensureblockspace}[1]{\par\begingroup\ifdim\pagegoal<\maxdimen\ifdim\dimexpr\pagegoal-\pagetotal\relax<#1\relax\newpage\fi\fi\endgroup}

\hypersetup{pdftitle={Threshold Compatibility and Support-Selection Diagnostics in Neural Discovery of Partial Differential Equations}}
\title{Threshold Compatibility and Support-Selection Diagnostics in Neural Discovery of Partial Differential Equations}
\author{Zeen Song {\small(\href{https://orcid.org/0009-0005-3552-4183}{0009-0005-3552-4183})}\\[0.4em]
\small School of Mathematics, Statistics and Mechanics\\
\small Beijing University of Technology, Beijing 100124, China\\
\small Correspondence: \href{mailto:zeensong17@gmail.com}{zeensong17@gmail.com}}
\date{17 October 2026}
\begin{document}
\maketitle
\section*{Abstract}\label{abstract}
\addcontentsline{toc}{section}{Abstract}

Accurate regression need not yield the correct equation after support selection. For a median/sample-standard-deviation threshold in neural PDE discovery, we derive an exact two-signal retention boundary and a complete-vector perturbation bound. Exact advection-diffusion libraries lose diffusion despite coefficient recovery to numerical precision. True-support refits of forty saved sparse states reduce diffusion error without restoring all derivative accuracy. A fixed sequential threshold ridge comparator recovers exact support in sixty ideal controls but selects extra terms in every learned state. All-fold correct-support candidates are rejected by conditional validation; post hoc energy pooling preserves their higher minimum loss in all sixty states. Supplied-support estimates outperform unchanged selected equations in all 240 physical-residual comparisons on four unused analytic trajectories. With coefficients, observation points and weights fixed, exact replacement of the learned library and target reverses their learned-row ordering in 59/60 states. Weighting both losses also fails the specified Burgers diffusion-benefit criterion at every tested budget. These finite diagnostics distinguish threshold incompatibility, refit damage and score rejection; they establish neither automatic correction nor general recovery or predictive stability.

\textbf{Keywords:} equation discovery; sparse regression; support selection; selective observation; neural approximation

\section{Introduction}\label{introduction}

Neural PDE discovery reconstructs a differentiable field, estimates coefficients and selects an equation \citep{both2021deepmod}. These operations can fail separately: a field may fit observations while its derivatives give a poor library, or a threshold may delete a coefficient that regression estimated accurately. We study whether such failures can be distinguished within one fixed pipeline.

Stage separation is established: sparse-sampling studies distinguish interpolation from estimation \citep{both2021sparse}, \emph{Sparsistent Model Discovery} examines analytic-library selection \citep{tod2021sparsistent}, DeepMoD discusses threshold sensitivity \citep{both2021deepmod}, and PINN-SR combines neural fields with sparse regression \citep{chen2021pinnsr}. Our STRidge comparator belongs to PDE-FIND \citep{rudy2016pdefind}. Weak-form alternatives include WSINDy's integrated systems and modified thresholding \citep{messenger2021wsindy} and WeakIdent's narrow fitting, contribution trimming and cross-validation \citep{tang2023weakident}; PDE-STRIDE uses subsampling stability \citep{maddu2022stride}. \emph{Freeze, Then Select} fits a field before weak-form selection \citep{zhong2026freeze}. We instead diagnose pointwise derivatives of a PDE-regularized field, establishing neither superiority nor general limits of these alternatives. Priority for the explicit boundary formula remains unestablished.

We link an exact threshold boundary and margin bound to support interventions, saved-path score rejection and evaluation of frozen equations. The study began with observation-inclusion weighting, whose negative Burgers result is retained. This weighting differs from importance sampling, residual-volume weighting and loss balancing \citep{horvitz1952generalization, nabian2021importance, song2024volume, heldmann2023biobjective}; its fixed-loss identity does not make a trained estimator unbiased.

\section{Threshold compatibility}\label{threshold-compatibility}

\subsection{Executed rule and assumptions}\label{executed-rule-and-assumptions}

Let \(A\in\mathbb R^{n\times p}\) be a fixed library, \(y\in\mathbb R^n\) a nonzero time-derivative target, and \(\beta\in\mathbb R^p\) a coefficient vector. Use the unweighted observed-row norms to define \(z_j=\beta_j\|A_j\|_2/\|y\|_2\). Columns used for normalized regression must have positive norm; uniqueness of its least-squares coefficient additionally requires full column rank. These regression assumptions are separate from the vector calculation below. Residual weights, when present, do not change the normalization convention.

Normalization and median centering follow DeepMoD (author supplement, Scaling and thresholding, equations (9)--(10)) \citep{both2021deepmod}. The lower median, sample SD and strict equality conventions come from the executed code; that paper does not specify them.

The executed support rule is

\[
\mathcal S(z)=\{j:|z_j-m(z)|>s(z)\},\qquad
s(z)^2=\frac{1}{p-1}\sum_{j=1}^p(z_j-\bar z)^2,
\]

where \(m\) is the lower median (the \(\lceil p/2\rceil\)-th order statistic), \(\bar z\) is the arithmetic mean, and \(p\ge2\). Equality is excluded. The experiments use \(p=12\).

\subsection{Exact boundary and stability}\label{exact-boundary-and-stability}

\textbf{Proposition 1 (two-signal retention).} Suppose \(p\ge4\) and \(z\) contains \(-a,b\) and \(p-2\) zeros, where \(0<b\le a\). The negative signal is retained, all zero entries are excluded, and the positive signal is retained if and only if

\[
\frac ba>r_p,\qquad
r_p=\frac{1+\sqrt{1+(p-1)^3}}{(p-1)^2}.
\]

\emph{Proof.} The lower median is zero and the mean is \((b-a)/p\). Hence

\[
s^2=\frac{(p-1)(a^2+b^2)+2ab}{p(p-1)}>0.
\]

For \(r=b/a\in(0,1]\), the ratio \(s^2/a^2\) increases with \(r\) and is at most \(2/(p-1)<1\). Thus \(a>s\) and zero coordinates are excluded. The condition \(b>s\) is equivalent to \((p-1)^2r^2-2r-(p-1)>0\). Its negative root is below zero and its positive root is \(r_p\). Equality gives exclusion because the selector is strict.

\ensureblockspace{1.1in}
\begin{samepage}
\textbf{Proposition 2 (margin stability).} For fixed \(p\ge2\) and arbitrary \(z,e\in\mathbb R^p\), write \(g_j(z)=|z_j-m(z)|-s(z)\). Then

\[
|g_j(z+e)-g_j(z)|\le
\left(2+\sqrt{\frac p{p-1}}\right)\|e\|_\infty.
\]
\end{samepage}

\emph{Proof.} Coordinatewise monotonicity and translation of order statistics give \(|m(z+e)-m(z)|\le\|e\|_\infty\), so the median-centered absolute coordinate changes by at most \(2\|e\|_\infty\). With \(v=(1,\ldots,1)^T\) and \(P=I-vv^T/p\), sample standard deviation is \(\|Pz\|_2/\sqrt{p-1}\). The reverse triangle inequality and orthogonal-projection bound give a change at most \(\|e\|_2/\sqrt{p-1}\le\sqrt{p/(p-1)}\|e\|_\infty\). Adding both estimates proves the result.

A negative margin remains negative if \(\|e\|_\infty<-g_j(z)/(2+\sqrt{p/(p-1)})\). This bounds perturbations of normalized vectors, not physical coefficients or derivatives separately. Proposition 1 does not apply to nonzero nuisance coefficients. At \(p=3\), equal opposite signals are both deleted at equality; the dimension assumption matters. Population variance or another median convention defines a different rule.

For the exact advection-diffusion equation \(u_t=D u_{xx}-v u_x\) with \(D,v>0\), define

\[
a=\frac{v\|u_x(O)\|_2}{\|u_t(O)\|_2},\qquad b=\frac{D\|u_{xx}(O)\|_2}{\|u_t(O)\|_2}.
\]

If these three norms are nonzero and \(b\le a\), the exact coefficient vector embedded in the library satisfies Proposition 1, with

\[
\frac ba=\frac Dv\frac{\|u_{xx}(O)\|_2}{\|u_x(O)\|_2}.
\]

The target norm cancels: retention depends on observed derivative geometry as well as physical coefficients. Positive residual weights preserve an exact unique coefficient vector and leave normalization unweighted; changing rows or spectrum can change the ratio. This gives no recovery guarantee for learned libraries.

For an exact regression witness, take \(A=I_{12}\), \(\beta=(-1,1/4,0,\ldots,0)\) and \(y=A\beta\). Every positive diagonal residual weight yields this unique coefficient vector. Normalization gives a common positive factor, yet \(s^2=65/704>1/16\) before scaling, so the weak signal is deleted. This is a regression example, not a PDE library. With zero-padded coefficients it is excluded at \(p=17\) and retained at \(p=18\): adding exact-zero terms can change selection. S1 adds a compensated column-sign example; S7 specifies formal coverage.

\section{Numerical design}\label{numerical-design}

\subsection{Common pipeline and repeated units}\label{common-pipeline-and-repeated-units}

For observed locations \(O\), \(n=|O|>0\), known inclusion probabilities \(\pi_i>0\) give mean-one weights \[
w_i=\frac{n/\pi_i}{\sum_{j\in O}1/\pi_j}.
\] With neural field \(f=f_\theta\), data errors \(\ell_i=(f(q_i)-u_i^{\rm obs})^2\) and PDE residuals \(r_i=f_t(q_i)-\Theta_i(f)\beta\), the ordered library is \[
\Theta(f)=(1,f_x,f_{xx},f_{xxx},f,ff_x,ff_{xx},ff_{xxx},f^2,f^2f_x,f^2f_{xx},f^2f_{xxx}).
\] Physical coefficients are \(\beta\). The penalty is \(P=\sum_{j\ne0}|\beta_j|\|\Theta_j(O)\|_2/\|f_t(O)\|_2\), with constant column indexed by zero and nonzero target norm required. The objectives are \[
\begin{aligned}
\mathcal J_{\rm M1}&=n^{-1}\sum(\ell_i+r_i^2)+\lambda P,\\
\mathcal J_{\rm data}&=n^{-1}\sum(w_i\ell_i+r_i^2)+\lambda P,\\
\mathcal J_{\rm both}&=n^{-1}\sum w_i(\ell_i+r_i^2)+\lambda P.
\end{aligned}
\] Sums use observed rows; the last two methods are H\_data and H\_both. Weights multiply squared residuals once, with no separate collocation set. Architecture, optimizer settings and evaluation metrics are specified in S3.

Both cases use 2,000 candidate points from a \(256\times101\) grid and Gaussian noise of scale 10\% of the clean field's standard deviation. Bernoulli masks depend only on location. The archived Burgers target is \(u_t=0.1u_{xx}-uu_x\) on \(x\in[-8,7.9375]\), \(t\in[0,10]\), with \(\pi(x)=0.1+0.9(x+8)/15.9375\). Its generator and boundary conditions remain unverified. The analytic case is \[
u(t,x)=0.2+\sum_{k=1}^3 a_k e^{-0.1k^2t}\cos\{k(x-0.7t)+\phi_k\},
\] on \([-\pi,\pi)\times[0,2]\), with \(a=(0.7,0.25,0.15)\), \(\phi=(0,0.4,-0.7)\) and \(\pi(x)=0.1+0.9(x+\pi)/(2\pi-2\pi/256)\). It satisfies \(u_t=0.1u_{xx}-0.7u_x\). Only noisy field values enter training; there is no boundary loss or supplied derivative truth.

Analytic blocks B1 and B2 use this same trajectory with different noise, initialization and ten masks per block; two placements give forty sparse states. B3 changes spectrum to \(a=(0.35,0.60,0.25)\), \(\phi=(0.5,-0.8,1.1)\) and changes random factors together, adding twenty sparse states but no neural refits. This is not a spectrum-only causal contrast. The Burgers loss-placement study has ten matched masks and three methods; four sparse/refit budget pairs reuse those masks. Seeds, nuisance controls and local freeze chronology are in S3 and S6. Masks within a block share noise and initialization; sixty analytic states represent two trajectories, not sixty independent problems.

\subsection{Support interventions and fixed-field diagnostics}\label{support-interventions-and-fixed-field-diagnostics}

Each B1/B2 sparse state branches into original-selected-support and supplied-true-support \(\{u_x,u_{xx}\}\) refits. Both copy the same field and retained learned coefficient values, reset Adam and take 25,001 updates. Truth supplies support, never coefficient values. Within each block/placement, the diffusion-error gain rule requires at least 7/10 improvements plus lower mean and median; aggregate field-MSE and observed-\(u_{xx}\) mean/median ratios must not exceed 1.10. All supports, adverse outcomes and ties remain in the denominator.

Separately, for frozen true-support design \(A\), learned target \(y\), full column rank and positive diagonal \(W\), define \[
b_W=(A^TWA)^{-1}A^TWy,\qquad L_W(b)=n^{-1}\|W^{1/2}(y-Ab)\|_2^2.
\] Normal equations give \[
\begin{aligned}
L_W(\widehat\beta)-L_W(b_W)&=n^{-1}\|W^{1/2}A(\widehat\beta-b_W)\|_2^2\ge0,\\
b_W-\beta_*&=(A^TWA)^{-1}A^TW(y-A\beta_*).
\end{aligned}
\] These oracle coefficient solves do not retrain the field. Signed optimization and physical errors can cancel; smaller residual does not imply smaller physical error. S2 gives the proof, derivative-error projection and counterexamples.

\subsection{Frozen-library selection and score rejection}\label{frozen-library-selection-and-score-rejection}

The STRidge comparator uses all sixty sparse states, five contiguous time folds and training-row-only norms. It is a fixed-grid diagnostic in the established algorithm family \citep{rudy2016pdefind}, not a reproduction of PDE-FIND tuning or its best achievable performance. The primary ridge is \(10^{-6}\) and the grid is \(\{0\}\cup\{10^{-6+j/8}:j=0,\ldots,48\}\). The loop deletes magnitudes strictly below threshold, retaining equality, then debiases without ridge or further thresholding. S4 specifies tie rules, weight normalization and degenerate inputs. Zero ridge and minimum-loss selection are prespecified sensitivities.

For fold \(V_k\) and its training-only coefficient estimate \(\widehat\beta_\tau^{(-k)}\), the executed score and acceptance ceiling are \[
\begin{aligned}
L_k(\tau)&=\frac{\sum_{i\in V_k}w_i(y_i-A_i\widehat\beta_\tau^{(-k)})^2}{\sum_{i\in V_k}w_i y_i^2},\qquad
\bar L(\tau)=\tfrac15\sum_{k=1}^5L_k(\tau),\\
C&=\bar L(\tau_*)+\mathrm{SE}(\tau_*)+10^{-12}.
\end{aligned}
\] Here \(A=\Theta(f)(O)\), \(y=f_t(O)\), \(\tau_*\) is the first grid minimizer, and SE is the sample standard deviation of the five losses divided by \(\sqrt5\). H\_data uses unit residual weights; H\_both uses the supplied inclusion weights. Among eligible thresholds, choose smallest mean fold support, then largest threshold, then smallest index. Only the chosen threshold is refitted on all observed rows. All sixty states have positive validation target energies.

For explanation only, let \(\mathcal T_{\rm true}\) contain thresholds with correct support in all five training folds and define \(\Delta=\min_{\tau\in\mathcal T_{\rm true}}\bar L(\tau)-C\) when this set is nonempty. Positive \(\Delta\) excludes every such candidate from eligibility; it says nothing about an uncomputed full-data refit. This explanation was post hoc in B1/B2 and prespecified in B3. The field has seen all observed rows before the folds are formed: the score is conditional on that representation, not independent end-to-end validation. One-SE is a sparsity heuristic, not a confidence interval.

\subsection{External physical scoring}\label{external-physical-scoring}

Four prescribed analytic trajectories, excluded from all representation training, use the same PDE with different Fourier amplitudes, phases and offsets (S5). Exact derivatives define their libraries and targets. All sixty selected and original median/SD equations remain unchanged. A supplied-support comparator fits coefficients only on each state's original learned rows and weights; it uses neither true coefficients nor external data for fitting. For trajectory \(v\), \[
E_v(\beta)=\frac{\int_0^2\int_{-\pi}^{\pi}[\partial_tu_v-\Theta(u_v)\beta]^2\,dx\,dt}{\int_0^2\int_{-\pi}^{\pi}(\partial_tu_v)^2\,dx\,dt},\qquad E(\beta)=\tfrac14\sum_{v=1}^4 E_v(\beta).
\] Reported scores are numerical quadrature approximations. All target energies are positive. Each block/placement requires 7/10 lower aggregate scores and lower mean and median, using strict tolerance \(10^{-6}\max(|a|,|b|,10^{-30})\). Spectra, rules and inputs were frozen locally before scoring, without external preregistration. No score tunes a threshold or coefficient. The \(60\times4=240\) comparisons reuse sixty fitted states; they are neither independent recovery trials nor forward simulations.

\section{Results}\label{results}

\subsection{Exact-data obstruction and matched support intervention}\label{exact-data-obstruction-and-matched-support-intervention}

All twenty original-spectrum exact-library solves have numerical rank twelve and recover physical coefficients within \(1.267\times10^{-15}\), yet median/SD removes diffusion in every case. Scaled diffusion/advection ratios 0.229026--0.241197 are below \(r_{12}=0.309889\) (Figure 1a); sufficient exclusion radii are 0.019412--0.023048 in normalized-vector infinity norm. Before selection, raw diffusion estimates range from 0.050861 to 0.105473, but all are deleted. Every original-refit field and derivative error worsens relative to its sparse start on observed rows (Figure 1).

\begin{figure}[!htbp]
\centering
\includegraphics[width=\linewidth,height=4.8in,keepaspectratio]{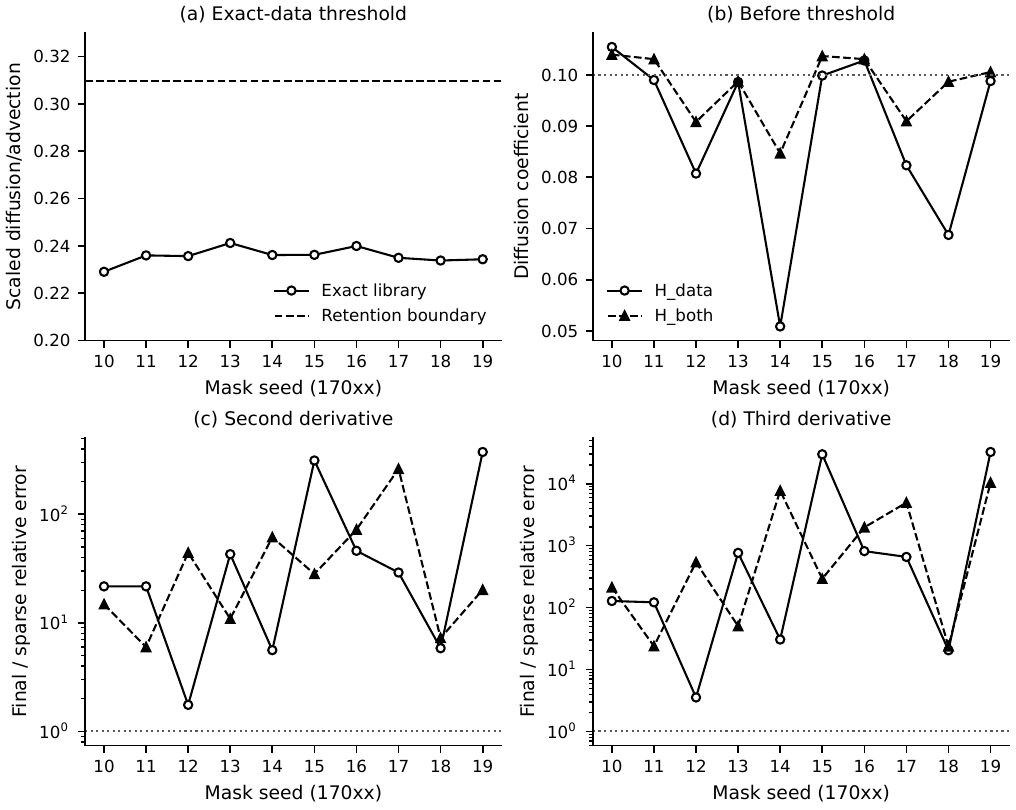}
\caption{Exact-data threshold qualification and saved phase endpoints. (a) All original-spectrum exact observed designs fall below the two-signal boundary; identity and IPW fits recover the same physical coefficients. (b) All twenty sparse-phase diffusion estimates before thresholding; the dashed line marks 0.1. (c,d) Final-refit/sparse observed derivative relative-error ratios after float64 reevaluation of saved float32 parameters; values above one indicate deterioration. These post hoc diagnostics change neither training nor support. Seed order is not time; all outcomes are retained}\label{fig:F10_threshold_qualification}
\end{figure}

The twenty original analytic fits therefore have 0/20 exact supports and diffusion error exactly 0.1. The ten diffusion ties do not show equivalence between placements. H\_both improves full-grid field MSE on 8/10 masks (mean 0.00136859 to 0.00111011), while mean grid relative-\(L^2\) \(u_{xx}\) errors remain 30.8901 and 18.1893 for H\_data/H\_both. Advection and full-vector coefficient errors improve on 6/10 pairs but their means increase. The float32/float64 evaluation screen fails for ten fits; reevaluation preserves the large errors and direction counts, without replacing float64 retraining (S3).

Supplying true support improves diffusion over original-support refits in all forty B1/B2 pairs and meets each block's mean/median and aggregate guardrails (Figure 2; Table S1). The baseline error 0.1 is imposed by omission; oracle errors remain nonzero. Field MSE improves in all B1 pairs and 9/10 and 10/10 B2 pairs. Seed 17115 has adverse H\_data grid MSE and H\_both grid derivatives; both placements worsen unobserved-candidate MSE there. Advection improves on only 4--8 of ten pairs per block/placement; complete counts are retained in the ancillary intervention tables.

\begin{figure}[!htbp]
\centering
\includegraphics[width=\linewidth,height=4.8in,keepaspectratio]{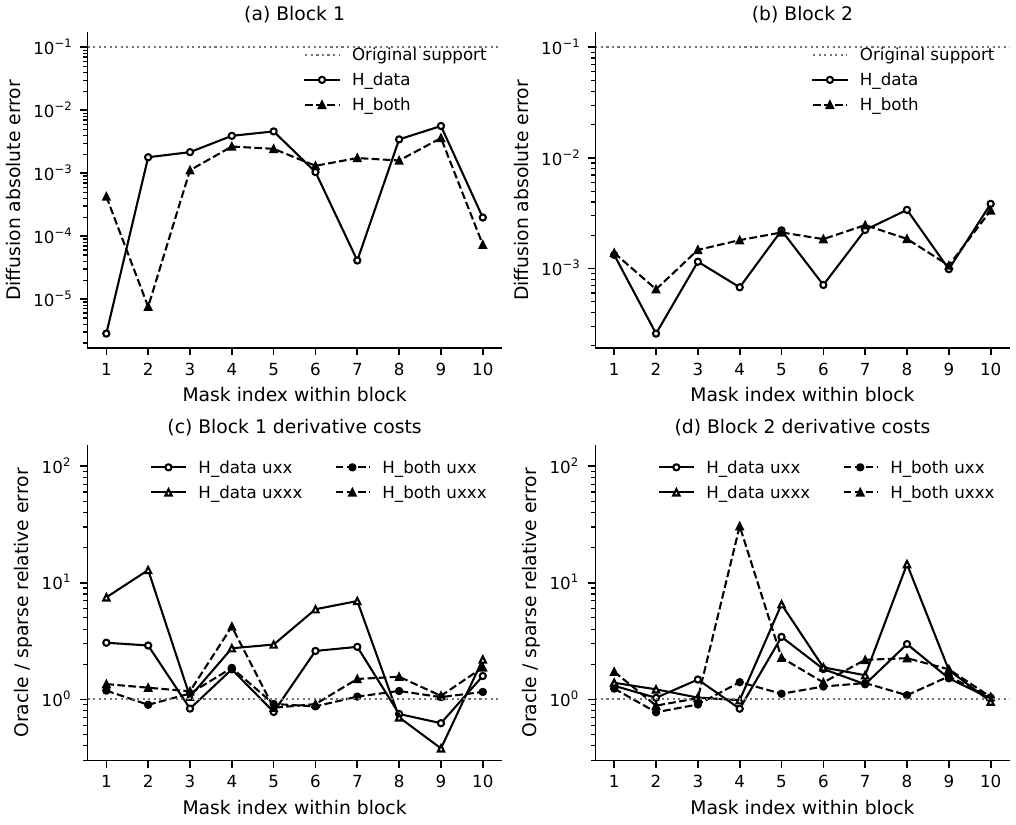}
\caption{True-support intervention and remaining derivative costs. Top: all forty oracle diffusion errors, compared with the original-support value 0.1. Bottom: oracle/sparse observed derivative relative-error ratios; values above one indicate deterioration. Circles denote $u_{xx}$ and triangles $u_{xxx}$. Mask indices 1--10 map to seeds 17010--17019 in B1 and 17110--17119 in B2. Both rows use logarithmic vertical axes. Supplied support is an oracle; retained coefficients are learned. All pairs remain, including adverse outcomes and precision flags; shared-condition sampling is described in Section 3.1}\label{fig:F11_support_intervention}
\end{figure}

Relative to sparse starts, oracle observed \(u_{xx}\) worsens on 6/10 and 7/10 B1 pairs and 9/10 and 8/10 B2 pairs; \(u_{xxx}\) worsens on 8/10 and 8/10, then 8/10 and 9/10. This contrast also removes the penalty and continues training, so it does not isolate a cause. The placements' mean oracle diffusion-error ranking reverses between blocks. Correct support mitigates omission damage without restoring all derivative accuracy.

\subsection{Conditional selection retains correct terms but selects extra ones}\label{conditional-selection-retains-correct-terms-but-selects-extra-ones}

STRidge recovers exact support in all 60 ideal controls, but in none of the 60 learned states (Figure 3). Every learned fit retains both true terms; B1/B2 cells average 6.6--7.4 extra terms and B3 averages 7.9 and 8.6. The old rule succeeds in 0/40 original-spectrum ideal controls and 2/20 new-spectrum controls: one mask under both weights. Its ratio 0.310951 exceeds \(r_{12}\); the other nine new-spectrum masks fall below it. This exception agrees with the formula and rules out unconditional failure of the old threshold.

\begin{figure}[!htbp]
\centering
\includegraphics[width=\linewidth,height=4.8in,keepaspectratio]{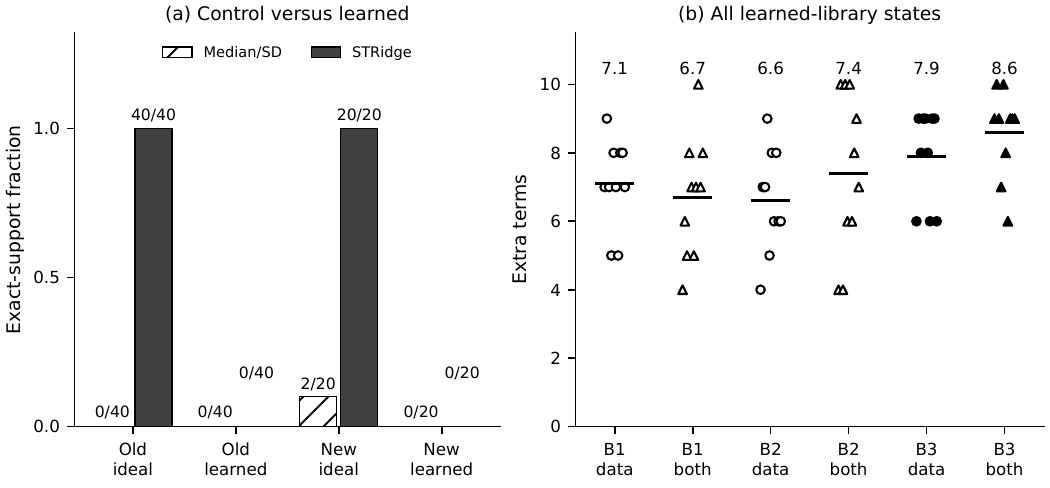}
\caption{Frozen selector comparison. (a) Exact-support fractions with denominators. Old denotes B1/B2; new denotes B3, which changes spectrum and random conditions together. Each ideal design uses two weights: the two new-spectrum median/SD successes are one mask under both measures. (b) All ten primary learned fits per placement/block, with mean marks and labels; each retains both true terms. Filled markers denote B3; circles/triangles denote $H_{\mathrm{data}}$/$H_{\mathrm{both}}$. Hatched/solid bars distinguish selectors. Counts reuse shared conditions (Section 3.1); validation is conditional on the fitted representation (Section 3.3)}\label{fig:F12_selector_comparison}
\end{figure}

Every learned state has \(\mathcal T_{\rm true}\ne\emptyset\) and \(\Delta>0\): the best all-fold correct-support loss is approximately 2.04--65.25 times the executed ceiling (Figure 4). Zero ridge and float64 reevaluation preserve primary supports; minimum-loss selection also yields no exact learned support. Post hoc pooling by validation target energy changes the minimizing threshold index in 5/60 states, but the best all-fold correct-support candidate still has higher loss than the unrestricted minimum in all sixty (S4). This last check concerns minimum loss, not a pooled one-SE rule. Error compensation is a possible explanation for extra terms, not an isolated causal finding.

\begin{figure}[!htbp]
\centering
\includegraphics[width=\linewidth,height=4.8in,keepaspectratio]{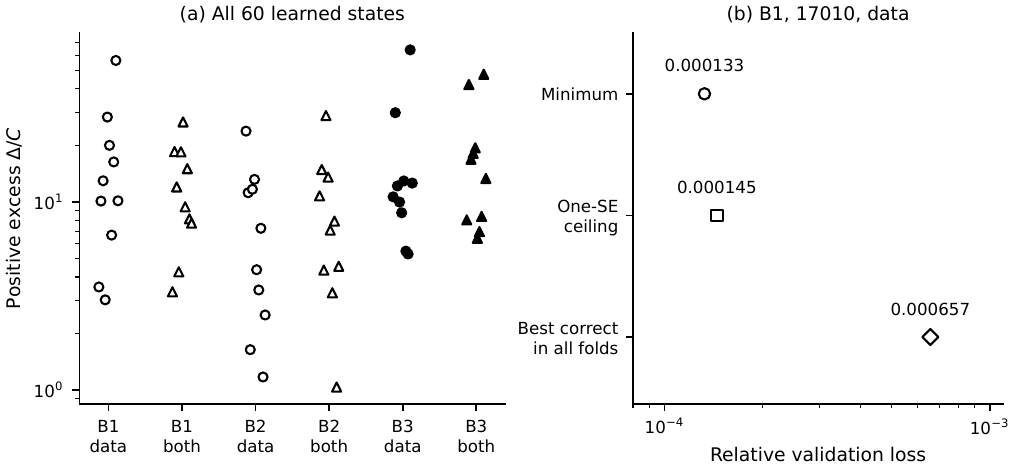}
\caption{Correct-support candidates rejected by the conditional score. (a) All sixty learned states; circles/triangles denote $H_{\mathrm{data}}$/$H_{\mathrm{both}}$, filled symbols denote B3, and offsets order masks. The logarithmic vertical axis shows positive excess $\Delta/C$: all candidates lie above the acceptance ceiling. (b) The first B1 data-only mask, chosen by identifier, shows minimum loss, the one-SE ceiling (including $10^{-12}$ slack), and the best threshold correct in all five training folds. Truth labels are diagnostic; unselected full-data refits were not computed. This heuristic gives no confidence interval}\label{fig:F13_selector_score_gap}
\end{figure}

\subsection{External scores reverse the learned-target ranking}\label{external-scores-reverse-the-learned-target-ranking}

Selected supports contain the supplied true support. Their original learned-row residuals are lower than those of the supplied-support fits, with selected/oracle ratios 0.008799--0.300751, consistent with nested least-squares spaces. On unused physical trajectories, every one of the 240 paired comparisons instead favors supplied support. Each block/placement wins 10/10 aggregate comparisons, with lower mean and median; the sixty aggregate selected/oracle ratios range from 2.39596 to 3356.22 (Figure 5). Selected equations still outperform the old median/SD equations in all 240 trajectory pairs. Absolute errors and all pairs are retained in the external CSV tables, so large ratios are not presented without their scale.

\begin{figure}[!htbp]
\centering
\includegraphics[width=\linewidth,height=4.8in,keepaspectratio]{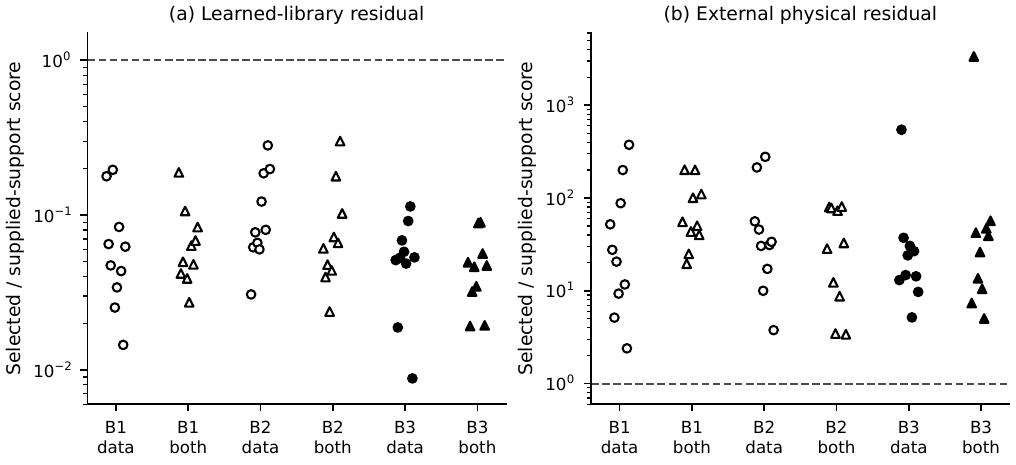}
\caption{Ranking reversal under external physical scoring. Each panel includes all sixty states grouped by block/placement, with within-cell offsets in mask order. Circles/triangles denote $H_{\mathrm{data}}$/$H_{\mathrm{both}}$; filled markers identify B3. Both panels use logarithmic vertical axes for selected/supplied-support squared-residual ratios; dashed lines mark equality. (a) Original learned rows and residual measure. (b) Equal mean of four unused-trajectory scores under uniform physical measure. Oracle support is supplied, but coefficients use only original learned rows. External data tune no model; these are correlated comparisons (Section 3.1), not independent recovery trials}\label{fig:F14_external_equation}
\end{figure}

A post hoc diagnostic holds observation coordinates, residual weights and all three coefficient vectors fixed. Replacing both the learned library and target by exact analytic values reverses the selected/supplied-support ranking in 59/60 states: 9/10 for B1 H\_data and 10/10 in each other cell, with no ties or invalid energies. The exception, seed 17018 in B1 H\_data, has selected/oracle ratios 0.014486 on learned rows, 0.936888 on same-point exact rows and 2.395964 externally. Thus a trajectory change is unnecessary for most reversals, but same-point replacement does not explain every external reversal. S5 gives crossed library/target scores and full state records. The supplied-support coefficients are the frozen-library estimates defined in Section 3.4, not joint neural-refit endpoints.

Quadrature refinement and float32/float64 evaluation preserve every direction (S5). Residual scoring alone cannot guarantee support identification: for \(u=e^{-Dt}\cos(x-vt)\), \(u_{xx}=-u\), so distinct supports \(\{u_x,u_{xx}\}\) and \(\{u_x,u\}\) give the same zero residual.

\subsection{Burgers weighting remains a negative result}\label{burgers-weighting-remains-a-negative-result}

In the Burgers loss-placement study, H\_both improves diffusion over H\_data on 4/10 masks, with higher mean (0.00155504 versus 0.00116425) and median (0.00120506 versus 0.00106527). All 30 supports are exact, but the S6 diffusion-benefit rule fails. Grid MSE improves on 3/10 masks; its mean rises from \(6.8628\times10^{-5}\) to \(1.14754\times10^{-4}\). Coefficient guardrails pass but exclude field error.

Across four sparse/refit budget pairs, diffusion wins are 4/10, 3/10, 5/10 and 2/10; none passes the rule. Every extended endpoint worsens grid MSE against its own original budget. Conditional fitting reduces residual without removing the adverse mean-diffusion contrast. These finite-budget comparisons reuse masks; nuisance controls and failed reconstructions remain in S6 and the ancillary tables.

\section{Discussion and conclusion}\label{discussion-and-conclusion}

The diagnostics distinguish weak-term deletion after exact regression, omission damage mitigated by supplied support despite derivative costs, and score rejection of available correct-support candidates. Pooling preserves the minimum-score discrepancy. Same-point exact replacement reproduces 59/60 external ranking reversals; the remaining state changes rank only with trajectory or evaluation measure in the tested comparisons. Their causal effects are not identified.

The scope is one archived Burgers trajectory, two analytic training trajectories, three shared condition blocks and one architecture/library/optimizer. Four external trajectories test the same PDE. There is no unweighted analytic neural baseline, isolated spectrum contrast, significance/equivalence test, noisy external study or forward integration. Truth supplies controls, support and external derivatives. Precision failures and adverse field/derivative outcomes remain; S6 records Burgers provenance limits.

The finite threshold obstruction differs from WSINDy's conditional asymptotic consistency \citep{messenger2025consistency}, which requires a specified weak system and sampling limit, with additional separation for full support recovery. It does not certify this learned pointwise library. Practical diagnostics should qualify selectors on exact identifiable designs, distinguish candidate absence from score rejection, and assess unchanged equations beyond the fitted representation. General recovery, neural convergence, architecture/noise robustness and predictive stability remain open; no deployable correction is supplied. S7 specifies formal coverage.

\section*{Statements and availability}\label{statements-and-availability}
\addcontentsline{toc}{section}{Statements and availability}

\textbf{Data and code availability.} Zeen Song (zeensong17@gmail.com) provides analytic rescoring and formal-verification attachments on reasonable request. They contain saved inputs, scripts, reference outputs and hashes, or Lean sources, mappings and logs (S8; \texttt{anc/README.txt}). They do not reproduce training. No public repository or full-pipeline replication is claimed; third-party Burgers data and full texts are excluded.

\textbf{Acknowledgments.} OpenAI Codex assisted with code, source inspection, literature comparison, drafting and computational and interpretive checks; these are not independent human review.

\textbf{Funding.} This research received no dedicated financial support.

\textbf{Compliance with ethical standards.} This study uses mathematical analysis, synthetic fields and archived numerical data, without human participants or animal experiments.

\textbf{Conflict of interest.} The author declares no financial or non-financial competing interests.

\textbf{Supplementary information.} S1--S8 follow the references; ancillary files contain full result tables, diagnostic CSVs and reproducibility notes.

\clearpage
{\small
\bibliographystyle{unsrtnat}
\bibliography{references}
}
\clearpage
\section*{Supplementary Material}
\setcounter{secnumdepth}{0}
\setcounter{table}{0}
\setcounter{figure}{0}
\renewcommand{\thetable}{S\arabic{table}}
\renewcommand{\thefigure}{S\arabic{figure}}
\renewcommand{\theHtable}{S\arabic{table}}
\renewcommand{\theHfigure}{S\arabic{figure}}
\section*{Reading guide}\label{reading-guide}
\addcontentsline{toc}{section}{Reading guide}

Sections S1--S6 provide derivations, execution details and adverse outcomes. S7 specifies formal coverage; S8 indexes result files and reproduction requirements.

\section{S1. Threshold conventions and boundary checks}\label{s1.-threshold-conventions-and-boundary-checks}

Use the normalization, lower median, sample SD and strict retention defined in Section 2. Proposition 1 requires \(p\ge4\) and \(0<b\le a\); at \(p=2\) the median need not vanish, and the \(p=3\) equal-signal failure is given there. Same-sign signals or population variance give different formulas. Learned vectors with nuisance coefficients require the full-vector margin calculation.

The recorded exact-arithmetic sweep of 6,100 dimension/ratio combinations checks code, not proof. Complete exact-design margins and phase endpoints remain in \texttt{anc/tables/S\_ORACLE\_exact.md} and \texttt{anc/tables/S\_ORACLE\_phases.md}; these display tables are not high-precision CSVs.

\textbf{Dependence on column orientation.} Take \(A=I_{12}\), \(\beta=(-1,3/10,0,\ldots,0)^T\) and \(y=A\beta\). For \(S=\operatorname{diag}(-1,1,\ldots,1)\), \(A'=AS\) and \(\beta'=S\beta\) give the same response and column norms. Both designs have unique exact fits under any positive diagonal residual weight. The vectors have lower median zero and unnormalized sample variances \(1259/13200\) and \(1139/13200\). Since \[
\frac{1139}{13200}<\left(\frac3{10}\right)^2=\frac{1188}{13200}<\frac{1259}{13200},
\] the weak term is deleted before the sign change and retained after it. Both SDs are positive and below one, so zeros are excluded and the strong term is retained. Normalization multiplies both vectors by \(10/\sqrt{109}\), preserving these strict comparisons. This finite regression witness does not show that changing column signs repairs the sampled PDE fits. Its proof/audit status is specified in S7.

\section{S2. Conditional fitting and analytic limitations}\label{s2.-conditional-fitting-and-analytic-limitations}

Use \(A,y,W,b_W,L_W\) from Section 3.2, with \(n,p>0\), full column rank and positive diagonal weights. Set \(G_W=A^TWA/n\). The result is finite and deterministic; no independence, population, differentiability or asymptotic assumption is needed.

For any nonzero \(h\), full column rank gives \(Ah\ne0\) and positivity gives \(h^TG_Wh=n^{-1}\sum_i W_{ii}(Ah)_i^2>0\). Thus \(G_W\) is invertible. The normal equations give \(A^TW(y-Ab_W)=0\). Expanding the quadratic around \(b_W\) therefore yields

\[
L_W(b_W+h)=L_W(b_W)+h^TG_Wh.
\]

This proves unique conditional optimality and the displayed residual-gap identity. Subtracting \(G_W\beta_*\) from \(G_Wb_W=A^TWy/n\) gives the remaining conditional physical-error formula. Adding and subtracting \(b_W\) gives the signed error decomposition. Multiplying \(W\) by a positive scalar scales both sides of the normal equations equally and preserves \(b_W\).

The assumptions matter. A rank-deficient \(A\) need not give a unique coefficient vector. Zero weights can remove rank, and negative weights need not give a convex quadratic. Our computations retain failures of these checks rather than interpreting an unstable inverse. The numerical implementation uses SVD rather than explicitly forming an inverse; the inverse notation describes the full-rank mathematical solution.

Residual improvement is distinct from physical accuracy. With \(n=p=1\), \(A=W=[1]\), \(y=1\) and \(\widehat\beta=\beta_*=0\), the fit is \(b_W=1\): residual loss decreases from one to zero, but coefficient error increases from zero to one. The signed decomposition \(0=(-1)+1\) shows why absolute component magnitudes cannot be added as causal percentages.

For the neural diagnostic, \(y-A\beta_*\) is a learned derivative/library mismatch, not an independently observed physical-derivative error. Conditional fitting quantifies a coefficient gap for the fixed field and weights; it does not certify joint stationarity or explain how the field was learned. S7 records formal coverage of this identity and the rank/weight counterexamples; numerical SVD code is outside that coverage.

\subsection{Analytic solution and derivative-error projection}\label{analytic-solution-and-derivative-error-projection}

For \(D>0\), real \(v,c,a_k,\phi_k\) and integer \(K>0\), write \(u=c+\sum_{k=1}^K b_k\cos z_k\), where \(z_k=k(x-vt)+\phi_k\) and \(b_k=a_ke^{-Dk^2t}\). Each term has spatial derivatives \(-kb_k\sin z_k\), \(-k^2b_k\cos z_k\), \(k^3b_k\sin z_k\) and time derivative \(-Dk^2b_k\cos z_k+vkb_k\sin z_k\). This finite sum is jointly smooth, spatially \(2\pi\) periodic and satisfies \(u_t=Du_{xx}-vu_x\). Restriction to \([0,T]\times\mathbb R\) for \(T\ge0\) uses the same derivatives, including endpoint derivatives through its global extension; no infinite-series interchange or numerical generator is involved.

At any fixed sample of size \(n\ge2\), suppose \(A=(f_x,f_{xx})\) has rank two and \(W\) is positive diagonal. The Gram \(A^TWA\) is positive definite. Define \(P_W=(A^TWA)^{-1}A^TW\) and \(\beta_*=(-v,D)^T\). Since \(P_WA=I\), the normal equations give

\[
\begin{aligned}
b_W-\beta_*&=P_W(f_t-A\beta_*)\\
&=P_W\{(f_t-u_t)+v(f_x-u_x)-D(f_{xx}-u_{xx})\}.
\end{aligned}
\]

Write \(e_t=f_t-u_t\), \(e_x=f_x-u_x\) and \(e_{xx}=f_{xx}-u_{xx}\). For \(B=W^{1/2}A\), full rank gives \(\sigma_{\min}(B)>0\) and \(\sigma_{\min}(B)\|h\|_2\le\|Bh\|_2\). Orthogonal weighted projection also gives \(\|W^{1/2}AP_Wd\|_2\le\|W^{1/2}d\|_2\). Applying both to \(d=e_t+ve_x-De_{xx}\) and then the triangle inequality yields

\[
\|b_W-\beta_*\|_2\le
\frac{\|W^{1/2}e_t\|_2+|v|\|W^{1/2}e_x\|_2+D\|W^{1/2}e_{xx}\|_2}
{\sigma_{\min}(B)}.
\]

The bound is invariant under positive scalar rescaling of \(W\). It can be arbitrarily weak near rank deficiency and does not order two learned fields from their derivative norms alone. The signed identity is used in the experiment; the bound is not claimed to be sharp or novel.

A finite example makes the projection limitation explicit. Let \(A\) have rows \((1,0),(0,1),(0,0)\) and \(W=I_3\). For any \(\beta_*\), mismatch vectors \(d_1=(0,0,2)^T\) and \(d_2=(0,1,0)^T\) have norms two and one, while their coefficient-error projections are \(A^Td_1=0\) and \(A^Td_2=(0,1)^T\). A smaller mismatch norm can therefore accompany a larger coefficient error. This is an exact finite regression example, not a claim about which arrays neural training produces.

The assumptions cannot be omitted. A constant field has zero derivative columns. A single Fourier mode satisfies \(u_{xxx}=-k^2u_x\) and \(u_{xx}=-k^2(u-c)\), so its full twelve-term library is rank deficient regardless of sample size. Three modes do not automatically prove a rank condition; every sampled design is checked separately. Positive weights and full rank give uniqueness only within the specified conditional support, not identification among arbitrary equations.

Small field error need not imply small derivative error: on \([0,2\pi]\), \(g_m(x)=\sin(mx)/m\) has \(\|g_m\|_2=\sqrt\pi/m\) and \(\|g_m''\|_2=m\sqrt\pi\). This exact limiting example does not describe the trained networks and, like the explicit full twelve-column rank argument above, is outside the formal coverage in S7.

\section{S3. Training protocol, support interventions and precision}\label{s3.-training-protocol-support-interventions-and-precision}

For a nonempty evaluation set \(\mathcal I\) (full grid, observed rows or unobserved candidate rows), define \[
\begin{aligned}
e_D&=|\widehat D-D|,\qquad
\operatorname{MSE}_{\mathcal I}=\frac1{|\mathcal I|}\sum_{i\in\mathcal I}(f_i-u_i)^2,\\
\operatorname{Rel}_{\mathcal I}(q)&=\frac{\|\widehat q-q\|_{2,\mathcal I}}{\|q\|_{2,\mathcal I}},
\qquad q\in\{u,u_t,u_x,u_{xx},u_{xxx}\}.
\end{aligned}
\] Here \(\widehat q\) is the field or derivative evaluated from the neural model, and \(q\) is analytic truth; the relative error requires a nonzero truth norm. These evaluation metrics are unweighted. Relative \(L^2\) errors are norm ratios, unlike the squared-residual scores in Sections 3.3--3.4. All analytic diffusion errors use \(D=0.1\).

Five hidden layers of 20 tanh units use float32 Adam at learning rate 0.002 and moments \((0.99,0.999)\). The sparse phase takes 25,001 updates at \(\lambda=10^{-5}\); after Section 2's threshold, field and retained coefficients are jointly refitted for 25,001 updates with fresh Adam and \(\lambda=0\).

All analytic blocks use candidate seed 17001, the same grid/library/propensity and two loss placements. B1 uses noise seed 17002, initialization 0 and masks 17010--17019; B2 uses noise 17102, initialization 1 and masks 17110--17119. B3 uses noise 17202, initialization 2 and masks 17210--17219, with the changed spectrum stated in the main text. Relative noise remains 10\% of field standard deviation, so its absolute amplitude changes in B3. Within each block the ten masks share noise and initialization; changing several factors together does not isolate their effects.

B1's twenty sparse states existed when the intervention protocol was fixed. The two refit branches follow Section 3.2. B2's protocol preceded its twenty sparse fits and forty refits; B3's spectrum and inherited selector preceded its twenty sparse fits. No seed replacement, outcome-driven budget extension or scientific retry is reported. Local freezes are not externally authenticated preregistration.

The intervention gate is defined in Section 3.2. Strict comparisons use \(10^{-6}\max(|a|,|b|,10^{-30})\); ties remain non-improvements. The original analytic loss-placement gate additionally required exact support in all twenty fits, unlike Burgers support safety. These distinct gates must not be pooled.

\begin{table}[!htbp]
\centering\small
\setlength{\tabcolsep}{4pt}
\caption{Matched true-support intervention in two separate condition blocks.}\label{tab:T_intervention_blocks}
\begin{tabular}{@{}llllll@{}}
\toprule
Block & Placement & Mean $e_D$ & Median $e_D$ & $e_D$ improved & $u_{xxx}$ error worsened \\
\midrule
1 & H\_data & 0.0022844 & 0.00197657 & 10/10 & 8/10 \\
1 & H\_both & 0.00149751 & 0.0014515 & 10/10 & 8/10 \\
2 & H\_data & 0.00167691 & 0.0012384 & 10/10 & 8/10 \\
2 & H\_both & 0.00179883 & 0.00182292 & 10/10 & 9/10 \\
\bottomrule
\end{tabular}
\par\smallskip\raggedright\footnotesize Original-support $e_D$ is 0.1 in every cell. $e_D$ improvements compare oracle with original-support refits; $u_{xxx}$ error deterioration compares oracle with sparse starts on observed rows. Each count has denominator ten. Masks share noise and initialization within a block; no pooling as independent random-condition replications.
\end{table}

Table S1 uses two different baselines deliberately: diffusion gains compare oracle and original-support refits, while derivative costs compare oracle with sparse starts. Complete endpoints, guardrails and phase comparisons remain in \texttt{S\_SI\_endpoints\_1/2.csv}, \texttt{S\_SI\_guards\_1/2.csv} and \texttt{S\_SI\_phase\_1/2.csv}. Besides the main-text adverse B2 seed 17115, B1 H\_data seed 17012 worsens grid \(u_{xx},u_{xxx}\) and seeds 17014 and 17018 worsen grid \(u_t\). Adverse metrics within a model are not independent failure samples.

Reload checks reproduce saved float32 fields, coefficients and arrays; separate derivative evaluation and conditional solves check arithmetic, not neural convergence. The analytic generator agrees with automatic differentiation within \(1.34\times10^{-15}\) and has sampled PDE residual at most \(3.34\times10^{-16}\). Numerical ranks are one for the constant full-library control, seven for the single-mode control and twelve for all ten original three-mode observed designs. Three modes do not by themselves prove rank at every sample.

The original twenty-fit precision screen compares float32/float64 evaluations relative to analytic truth norms against \(10^{-3}\). Ten fits fail. Maximum discrepancies are 0.001803 for \(u_{xx}\) and 0.68235 for \(u_{xxx}\); all paired field/derivative directions persist in double evaluation. This converts saved float32 parameters and checks arithmetic/coordinate rounding, not training in float64. The failed screen is retained as PASS WITH PRECISION CAVEAT.

\ensureblockspace{2.3in}
\begingroup
\small
\setlength{\tabcolsep}{3pt}\renewcommand{\arraystretch}{1.14}
\begin{longtable}{@{}>{\raggedright\arraybackslash}p{0.5in}>{\raggedright\arraybackslash}p{1.03in}>{\raggedright\arraybackslash}p{1.03in}>{\raggedright\arraybackslash}p{1.1in}>{\raggedright\arraybackslash}p{1.1in}>{\raggedright\arraybackslash}p{0.43in}@{}}
\caption{Float64 reevaluation of saved parameters: both precision scales.}\label{tab:S_AD_precision}\\
\toprule
Field & Max diff. / truth & Max diff. / learned & Data mean error & Both mean error & Wins \\
\midrule
\endfirsthead
\multicolumn{6}{l}{Continued}\\
\toprule
Field & Max diff. / truth & Max diff. / learned & Data mean error & Both mean error & Wins \\
\midrule
\endhead
\bottomrule\endfoot
$u$ & 4.32749e-07 & 4.32777e-07 & 0.0723199 & 0.065128 & 8/10 \\
$u_t$ & 9.74597e-06 & 7.36233e-06 & 0.710358 & 0.590171 & 7/10 \\
$u_x$ & 1.03922e-05 & 6.90312e-06 & 0.569706 & 0.449637 & 6/10 \\
$u_{xx}$ & 0.00180266 & 1.10373e-05 & 30.8901 & 18.1893 & 6/10 \\
$u_{xxx}$ & 0.682347 & 2.50027e-05 & 3867.31 & 1160.75 & 6/10 \\
\end{longtable}
\noindent\footnotesize Maxima are over twenty fits. Mean errors are reevaluated grid relative L2 errors against analytic truth. Wins favor H\_both; all counts match the original float32 evaluation. This is reevaluation, not retraining.\par
\endgroup

The forty original true-support conditional solves have rank two; SVD versus separate QR differs by at most \(1.23\times10^{-15}\), and the signed projection identity closes within \(1.60\times10^{-15}\). In general, zero-filling only true-support coefficients is not equivalent to evaluating a model containing false positives; the original selected supports here are subsets of truth. Intervention precision screens flag ten original-support B1 models and nine B2 models but zero oracle refits. Each block's 200 paired field/derivative directions agree across evaluation precisions. Numerical agreement does not certify floating-point training or joint optimization.

\section{S4. Exact selection protocol and path evidence}\label{s4.-exact-selection-protocol-and-path-evidence}

The grid, threshold/debiasing rule, score, ceiling and tie order are defined in Section 3.3. Implementation normalizes columns and target by unweighted training-row norms, renormalizes positive training weights to mean one, and applies ridge to every active normalized coefficient, including the constant. Fits use float64 augmented least squares, with physical units restored after debiasing.

Zero columns are excluded; a zero training target returns empty support, and an empty selected model is allowed. NumPy array\_split partitions distinct observed times into five contiguous groups, keeping equal times together. For the Section 3.3 score, a zero denominator gives zero if its numerator is zero and \(10^{300}\) otherwise. No primary learned state invokes this fallback: all 300 validation target energies are positive, ranging from 24.4621 to 260.2861.

Original-rule baselines apply the Section 2 selector to trained or full-library conditional coefficients. Selected estimates are fixed-library fits, not neural refit endpoints. The ideal controls comprise thirty distinct observed designs under two residual measures.

Table S3 gives the complete per-cell gap ranges; individual candidate paths remain in the local executable archive. The path explanation was post hoc in B1/B2 and fixed prospectively for B3.

\begin{table}[!htbp]
\centering\small
\setlength{\tabcolsep}{4pt}
\caption{Conditional score rejection of all-fold correct-support candidates.}\label{tab:S_score_gap}
\begin{tabular}{@{}llllll@{}}
\toprule
Block & Placement & Rejected & Min. excess & Max. excess & Min. candidates \\
\midrule
1 & H\_data & 10/10 & 3.02195 & 56.4414 & 1 \\
1 & H\_both & 10/10 & 3.33454 & 26.6163 & 4 \\
2 & H\_data & 10/10 & 1.17593 & 23.7894 & 4 \\
2 & H\_both & 10/10 & 1.03995 & 28.7687 & 4 \\
3 & H\_data & 10/10 & 5.29253 & 64.2488 & 5 \\
3 & H\_both & 10/10 & 6.40998 & 47.6414 & 7 \\
\bottomrule
\end{tabular}
\par\smallskip\raggedright\footnotesize Excess is (best all-fold correct-support loss minus the executed one-SE ceiling) divided by that ceiling. All values are positive. Min. candidates is the smallest number of all-fold correct-support thresholds over the ten states. These truth-labeled diagnostics include block 3 for comparison; all 60 raw values and source paths are in selector\_score\_gaps.csv. No unselected full-data support is asserted.
\end{table}

Historical reductions checked all 15,000 primary validation losses from saved matrices and coefficients, with maximum discrepancy \(2.78\times10^{-17}\) and unchanged choices/gap signs; separate QR audits agreed with the comparator. These checks reused stored fits. The new aggregation check below separately rescores the saved paths, without rerunning coefficient fits.

Equal fold averaging is material: target energies \((1,1,1,1,100)\) and residual energies \((0,0,0,0,100)\) versus \((1,1,1,1,0)\) give mean ratios \(1/5\) versus \(4/5\), but pooled ratios \(25/26\) versus \(1/26\). This exact constructed ranking reversal motivates a saved-data sensitivity check. Mean fold support also need not equal support after full-data refitting; the conditional-validation limitation is stated in Section 3.3.

\textbf{Post hoc pooled-score sensitivity.} Let \(D_k=\sum_{i\in V_k}w_i y_i^2>0\) and use the fold losses from Section 3.3. Define \[
L_{\rm pool}(\tau)=\frac{\sum_{k=1}^5 D_kL_k(\tau)}{\sum_{k=1}^5D_k}.
\] Reusing the five-fold coefficient paths of all sixty learned float64 states at ridge \(10^{-6}\) changes the first minimizing threshold index in 5/60 states. In all sixty, \(\min_{\tau\in\mathcal T_{\rm true}}L_{\rm pool}(\tau)-\min_\tau L_{\rm pool}(\tau)>0\), with gaps \(1.10950\times10^{-4}\)--\(1.10248\times10^{-2}\). All 3,000 candidate scores and 15,000 fold losses were checked. Separate summation agrees within \(4.17\times10^{-17}\) for candidate scores; direct scalar rescoring of sixty minimizers and sixty best all-fold correct-support candidates agrees within \(8.68\times10^{-19}\). S8 indexes the retained outputs. No field or coefficient was fitted. This is a minimum-loss sensitivity, with no pooled one-SE rule or conclusion about uncomputed full-data refit supports.

\section{S5. External fields, controls and interpretation}\label{s5.-external-fields-controls-and-interpretation}

For \(k=1,\ldots,4\), the unused trajectories are \(u=\mu+\sum a_k e^{-0.1k^2t}\cos[k(x-0.7t)+\phi_k]\) on the same domain:

V1: \(\mu=0.2\), \(a=(0.55,0.20,0.10,0.05)\), \(\phi=(0.2,-0.4,0.9,-0.3)\).

V2: \(\mu=0.2\), \(a=(0.20,0.25,0.40,0.15)\), \(\phi=(-0.6,0.3,-1.0,0.7)\).

V3: \(\mu=-0.15\), \(a=(0.70,0.25,0.15,0)\), \(\phi=(1.2,-0.3,0.8,0)\).

V4: \(\mu=0.5\), \(a=(0.25,0.15,0.10,0.45)\), \(\phi=(0.9,0.1,-0.5,-1.2)\).

These prescribed stress cases vary several factors of one PDE jointly. Their protocol, source hashes and gates preceded scoring. The supplied-support solve defined in Section 3.4 need not equal an uncomputed all-row refit of a discarded threshold.

Uniform spatial quadrature uses 128 periodic nodes; time uses 64 Gauss--Legendre nodes on \([0,2]\). Refinement doubles both. Four relative integrals are averaged equally, not pooled by target energy. Exact-equation scores are below \(10^{-24}\) and zero-equation scores equal one within \(10^{-12}\). Primary/refined scores differ by at most \(5.14\times10^{-16}\), within the frozen \(10^{-12}+10^{-8}|E_{\rm refined}|\) bound. Float32 evaluation passes \(10^{-7}+10^{-3}|E_{64}|\) and preserves every aggregate and trajectory-specific direction.

All 24 block/placement/trajectory summaries are in \texttt{S\_external\_trajectories.csv}; all sixty aggregate scores for each of the three methods are in \texttt{S\_external\_states\_1/2/3.csv}. Every denominator and adverse original-selection outcome is retained. Section 4.3 gives the single-mode identifiability counterexample.

\subsection{Same-point replacement on the original observations (post hoc)}\label{same-point-replacement-on-the-original-observations-post-hoc}

Reuse the sixty B1/B2/B3 sparse states and the external study's unrounded selected, frozen-library supplied-support (\texttt{oracle\_support}) and median/SD coefficient vectors. All masks, row orders, coordinates and weights match the observation archives by hash. The coefficient estimates remain fixed; their distinction from joint neural refits is stated in Section 4.3.

Let \(A_L,y_L\) denote the stored float64 library/target, originally obtained by float64 evaluation of saved float32 neural parameters. This revision uses those arrays directly. At the identical coordinates, \(A_E,y_E\) are rebuilt from the three-mode analytic solution, replacing every factor in nonlinear library columns. The observed noisy field values are retained for provenance and Figure S1(a); they are not a time-derivative target. Residual weights are \(W=I\) for H\_data and mean-one inverse inclusion weights for H\_both. The archive stores unnormalized \(1/\pi_i\): dividing once by its observed mean leaves normalized scores and rankings invariant, but fixes the reported SSE and SSE\(/n\) convention. This discrepancy with the handoff's storage description is recorded explicitly.

For \(a,b\in\{L,E\}\), define \[
D_L=y_L^TWy_L,\quad D_E=y_E^TWy_E,\quad
Q_{ab}(\beta)=\frac{(y_b-A_a\beta)^TW(y_b-A_a\beta)}{D_E}.
\] The original-scale learned score is \(R_L=Q_{LL}D_E/D_L\) and the exact same-point score is \(R_E=Q_{EE}\). All four \(Q\) values use the same true denominator; LE and EL are algebraic mixed-source diagnostics, not complete physical fields. Every raw SSE, SSE\(/n\), both target energies and input hash are saved. Nonpositive energies would be retained as invalid without adding an epsilon; none occurred. Pairwise ties use \(10^{-6}\max(|q_1|,|q_2|,10^{-30})\) and are not wins. The six groups are descriptive; no independence-based significance test or new 7/10 success gate is used.

\begin{table}[!htbp]
\centering\small\setlength{\tabcolsep}{3pt}
\caption{Fixed-equation same-point diagnostic. Entries are medians of normalized squared-residual scores for selected (S) and supplied-support (O), computed separately across ten states. Each S; O entry gives two scores, not their ratio. External scores first average four trajectory risks within each state. Reversals compare LL with EE. All six cells have zero ties and invalid states.}\label{tab:samepoint}
\begin{tabular}{lrrrr}\toprule
Block / placement & Reversals & Learned: S; O & Exact: S; O & External: S; O \\\midrule
B1 / data & 9/10 & 6.98e-05; 0.00105 & 0.000885; 5.97e-05 & 0.00415; 7.71e-05 \\
B1 / both & 10/10 & 5.75e-05; 0.00114 & 0.00104; 4.85e-05 & 0.00341; 5.91e-05 \\
B2 / data & 10/10 & 4.96e-05; 0.000565 & 0.000535; 7.98e-05 & 0.00256; 0.000136 \\
B2 / both & 10/10 & 3.82e-05; 0.000646 & 0.000647; 0.000144 & 0.00369; 0.000156 \\
B3 / data & 10/10 & 2.71e-05; 0.000394 & 0.000527; 0.000148 & 0.00423; 0.000176 \\
B3 / both & 10/10 & 2e-05; 0.000431 & 0.000551; 0.00011 & 0.00373; 0.000114 \\
\bottomrule\end{tabular}
\end{table}

Table S4 gives the 59/60 supplied-support wins. The B1 H\_data exception, seed 17018, has exact selected/supplied-support scores 0.00453556/0.00484109, ratio 0.936888. Its supplied-support diffusion estimate is approximately 0.06878661 rather than 0.1, so exact scoring does not make this comparator a zero-error reference. LE favors selected in 16/60 states and EL in 46/60, versus 60/60 for LL and 1/60 for EE, with no ties. Both sources affect rankings; this algebra does not identify a unique causal allocation. CSVs retain all median-model comparisons and adverse orderings. The 240 external comparisons reuse the same equation sets.

\begin{figure}[!htbp]
\centering
\includegraphics[width=\linewidth,height=4.8in,keepaspectratio]{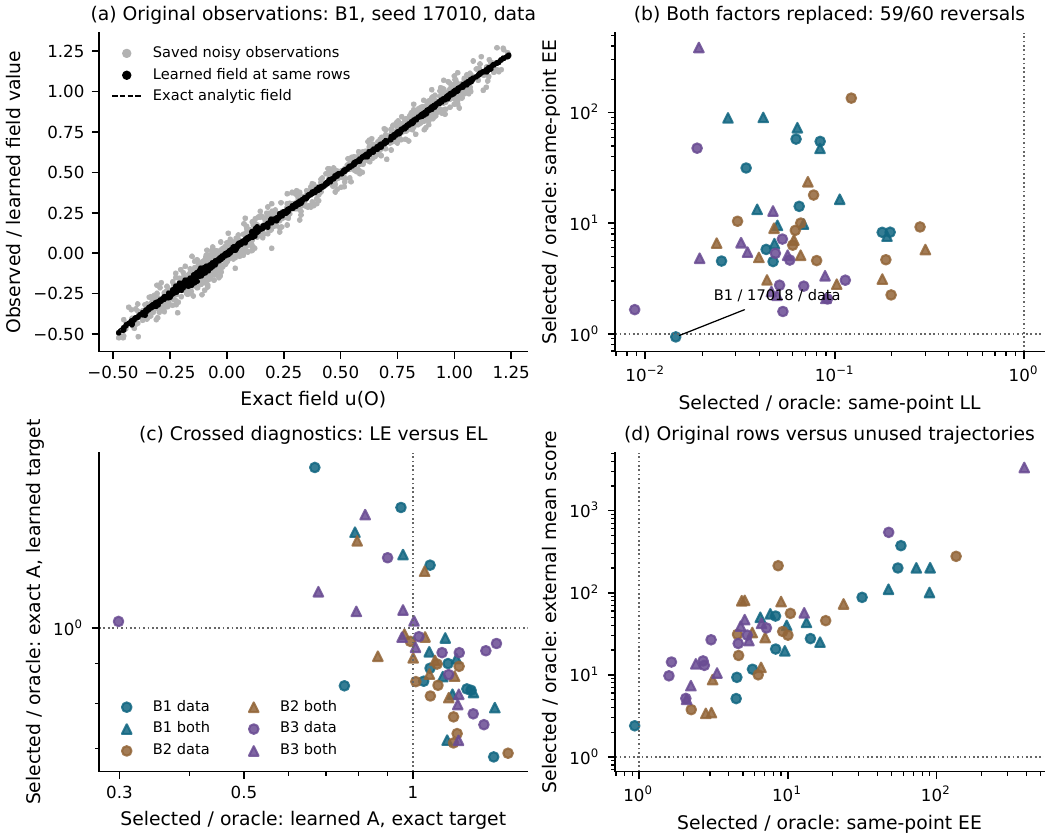}
\caption{Post hoc fixed-equation diagnostic. (a) All noisy observations and learned field values for the first B1 $H_{\mathrm{data}}$ state, chosen by identifier, against exact values at those rows; these are field measurements, not derivative targets. (b) All sixty selected/supplied-support ratios on learned LL versus exact EE rows, with coordinates/weights fixed. (c) Mixed-source LE versus EL. (d) EE versus the equal mean of four unused-trajectory scores. In (b)--(d), axes are logarithmic, dotted lines mark equality, and ratios below one favor selected. Colors denote blocks; circles/triangles denote $H_{\mathrm{data}}$/$H_{\mathrm{both}}$. All states remain, including the labeled B1 seed-17018 exception. CSVs retain all three estimators and absolute scores; these residuals are not forward prediction errors}\label{fig:F15_same_point_diagnostic}
\end{figure}

Independent complex-exponential, QR and scalar-summation checks agree within recorded tolerances for external and same-point scores. Environment replays preserve every ranking but do not establish cross-platform byte identity. These historical audits and their numerical discrepancies are retained in \path{anc/REPRODUCIBILITY_NOTES.txt}; no coefficient fit or neural update was rerun for this revision.

\section{S6. Burgers evidence, historical failures and data provenance}\label{s6.-burgers-evidence-historical-failures-and-data-provenance}

The loss-placement study (archive label WL) uses masks 7626--7635, noise seed 42 and initialization 0; all methods share observations. Its diffusion rule requires 7/10 improvements, lower mean and median and no loss of exact support. Four coefficient guardrails limit combined/advection mean/median ratios to 1.10, without covering field MSE. Combined error is \([(\widehat\beta_d-0.1)^2+(\widehat\beta_a+1)^2]^{1/2}/\sqrt{1.01}\); spurious magnitudes are excluded and support is assessed separately. None of the 74 original loss-placement and noise/initialization fits was recorded as meeting the sparse-phase gradient tolerance; they are not certified converged solutions.

Sparse/refit pairs are low/low, low/high, high/low and high/high (25,001/50,002 updates for low/high). Twenty branched trajectories produce eighty endpoints on the same ten masks. All four budget pairs keep exact support but fail diffusion benefit. Grid-MSE wins are 3/10, 3/10, 5/10 and 7/10, yet every extended endpoint worsens against its own original budget. Conditional fitting preserves the adverse H\_both mean-diffusion contrast, although some medians reverse. Original-budget replay is not an independent replicate.

The noise/initialization study (RB; five reused geometries) has exact support in all 44 fits, but its coefficient--field discordance rule holds in only 3/4 cells and its advection--diffusion branch in 1/4. Earlier exploratory blocks pass/fail their combined-error gates at 7/10 and 6/10; post-confirmatory pooling cannot reverse the failed decision. Diffusion improves in only 3/10 masks in each. These results establish neither uniform diffusion harm nor an error-amplification mechanism. Full designs, historical labels and reference models are in \texttt{anc/METHODS\_AND\_PROVENANCE.txt}.

Truth-tuned local-polynomial field repair gives 0/10 candidate-field improvements of 5\% and 1/10 in the low-propensity unobserved subset in each of two blocks, with worse support. Its failure does not test successful repair. The required diffusion gain 0.005 exceeds the maximum baseline error 0.0036769837 and is unattainable; a kernel candidate has no fitted performance. All failed outcomes remain in the historical tables.

The archived array is PhIMaL/DeePyMoD\_torch's \texttt{examples/data/burgers.npy}, numerically matching PDE-FIND's \texttt{Datasets/burgers.mat}. Commit hashes, file identity, initial-snapshot and imaginary-part checks are retained in \texttt{anc/METHODS\_AND\_PROVENANCE.txt}. The target coefficient remains 0.1 despite an inconsistent upstream supplement row \citep{rudyPdeFindSupplement}. The generator, boundary conditions and data redistribution remain unverified. The restored pipeline is not byte-level reproduction of the original publication; code licensing does not resolve raw-data rights. Third-party raw data are excluded.

\section{S7. Formal-verification scope}\label{s7.-formal-verification-scope}

Lean verification covers finite weighting identities; conditional fitting and rank/weight counterexamples (T-0014); Fourier/PDE identities and projections (T-0015); threshold/margin results (T-0016); and the compensated column-sign witness (T-0017). The last includes full rank, unit column norms, arbitrary positive residual weights, unique exact fits and complete normalized-fit supports; it is not a PDE-library construction.

On 17 October 2026, fifteen R1 modules and 204 theorem roots, including twenty-nine for T-0017, passed checks using only standard Lean axioms. The separate archive retains sources, pinned dependency identities, mappings and logs. Coverage excludes the field-\(L^2\) limiting example, the full twelve-column PDE-library rank argument, numerical solvers, training and empirical outcomes. Semantic audits are by the assistant, without human proof review. Pooled scores remain numerical evidence; no general recovery, convergence or priority claim is made.

\section{S8. Ancillary files and reproducibility access}\label{s8.-ancillary-files-and-reproducibility-access}

\texttt{anc/README.txt} indexes complete historical tables, same-point/crossed scores, pooled candidate/state results, methods, provenance and verification records. \texttt{anc/TABLE\_INDEX.csv} maps historical table identifiers to this paper.

Fixed-equation rescoring, complete pooled-path replay and formal-verification attachments are available on reasonable request. Their READMEs specify inputs, commands, hashes, dependencies and scope. They exclude neural training and third-party Burgers raw data; full training archives remain separate.

\end{document}